\documentclass[letterpaper, 10 pt, conference]{ieeeconf}  
\usepackage{amsmath,amsfonts}
\usepackage{algorithmic}
\usepackage{algorithm}
\usepackage{array}
\usepackage[caption=false,font=normalsize,labelfont=sf,textfont=sf]{subfig}
\usepackage{textcomp}
\usepackage{stfloats}
\usepackage{url}
\usepackage{verbatim}
\usepackage{graphicx}
\usepackage{cite}
\usepackage{booktabs}
\usepackage{makecell}
\usepackage{multirow}
\usepackage{float}
\usepackage{bbding}
\usepackage{wrapfig,lipsum}
\usepackage{subfig}
\usepackage{mathtools}

\IEEEoverridecommandlockouts                              

\title{\LARGE \bf
POD: Predictive Object Detection with Single-Frame FMCW LiDAR Point Cloud
}

\author{Yining Shi$^{1,2}$, Kun Jiang$^{1,2\dagger}$, Xin Zhao$^{1,2}$, Kangan Qian$^{1,2}$, \\Chuchu Xie$^{1,2}$, Tuopu Wen$^{1,2}$, Mengmeng Yang$^{1,2}$, Diange Yang$^{1,2\dagger}$
\thanks{All authors are with $^{1}$ School of Vehicle and Mobility, Tsinghua University, Beijing, China, and $^{2}$ State Key Laboratory of Intelligent Green Vehicle and Mobility, Beijing, China. $\dagger$ Corresponding authors: Diange Yang, Kun Jiang (ydg@mail.tsinghua.edu.cn, jiangkun@mail.tsinghua.edu.cn.)}
}

\begin{document}

\maketitle
\thispagestyle{empty}
\pagestyle{empty}

\begin{abstract}
LiDAR-based 3D object detection is a fundamental task in the field of autonomous driving. This paper explores the unique advantage of Frequency Modulated Continuous Wave (FMCW) LiDAR in autonomous perception. Given a single frame FMCW point cloud with radial velocity measurements, we expect that our object detector can detect the short-term future locations of objects using only the current frame sensor data and demonstrate a fast ability to respond to intermediate danger. To achieve this, we extend the standard object detection task to a novel task named predictive object detection (POD), which aims to predict the short-term future location and dimensions of objects based solely on current observations. Typically, a motion prediction task requires historical sensor information to process the temporal contexts of each object, while our detector's avoidance of multi-frame historical information enables a much faster response time to potential dangers. The core advantage of FMCW LiDAR lies in the radial velocity associated with every reflected point. We propose a novel POD framework, the core idea of which is to generate a virtual future point using a ray casting mechanism, create virtual two-frame point clouds with the current and virtual future frames, and encode these two-frame voxel features with a sparse 4D encoder. Subsequently, the 4D voxel features are separated by temporal indices and remapped into two Bird's Eye View (BEV) features: one decoded for standard current frame object detection and the other for future predictive object detection. For the feature encoding of 4D virtual points, we extend two mainstream voxel encoders, Sparse Convolutional (SparseConv) VoxelNet and Voxel Transformer, to 4D SparseConv VoxelNet and 4D Voxel Transformer. We compare the accuracy and latency of the two 4D encoder networks. Extensive experiments on our in-house dataset demonstrate the state-of-the-art standard and predictive detection performance of the proposed POD framework.
\end{abstract}


\section{Introduction}
Light Detection And Ranging (LiDAR) is one of the important sensors for autonomous vehicles. LiDAR-based 3D object detection can accurately detect the bounding boxes of typical objects. There are many variants in the hardware development of LiDAR, and perception algorithms need to be designed for specific hardware. Most works are designed for traditional 3D LiDARs, including mechanically rotating LiDARs and solid-state LiDARs, which typically return 3D point clouds and reflection intensities. VeloVox\cite{velovox} uses a Livox LiDAR with a non-repetitive scanning pattern with timestamps associated with point clouds. Therefore, VeloVox\cite{velovox} can extract the spatio-temporal features of a single-frame point cloud, and obtain the speed of each detected object.

\begin{figure}[ht]
	\centering
\includegraphics[width=0.45\textwidth]{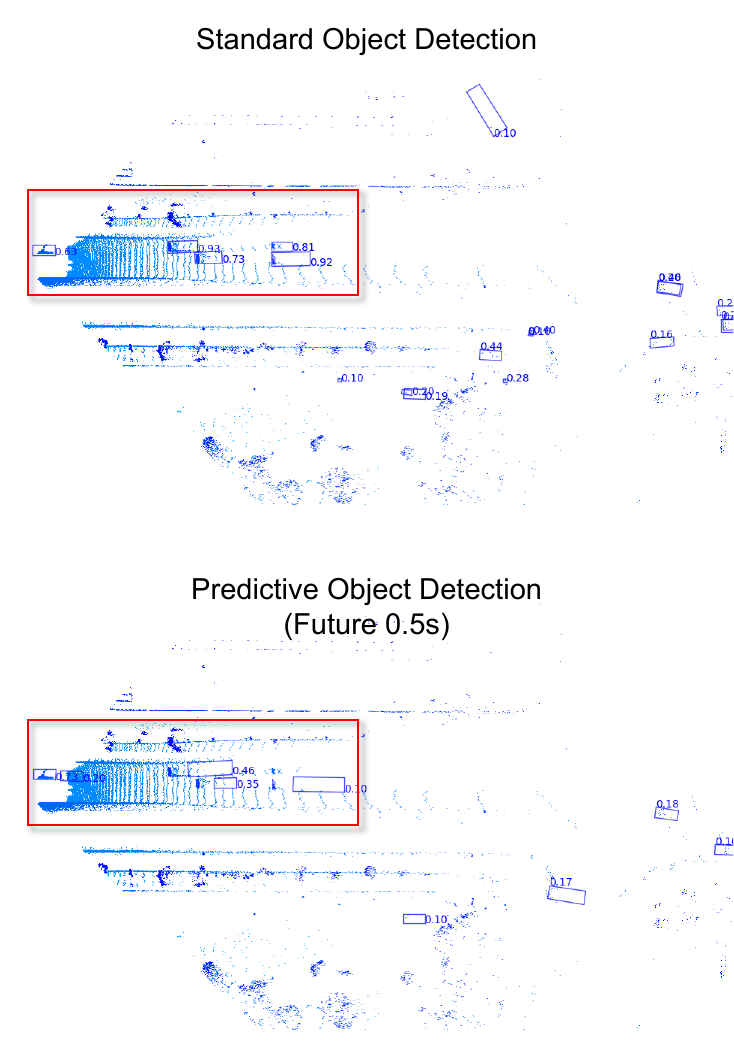}
	\caption{An one-frame example of the formulations of standard object detection and predictive object detection. Areas ahead of the ego vehicle is highlighted by red boxes.
	}
	\label{fig:POD_intro}
\end{figure}

This paper focuses on the perception algorithms of Frequency Modulated Continuous Wave (FMCW) LiDAR, also known as 4D LiDAR. FMCW LiDAR features the ability to measure radial velocity with the Doppler effect while scanning ambient point clouds. We explore whether there would be a unique advantage of using LiDAR with a velocity indicator for autonomous perception tasks. 

We start by using well-established detection networks with velocity information as input scalars. The raw relative velocity is compensated to the absolute velocity by clustering the ground points and compensating the mean velocity of grounded points as the opposite number of ego velocity. We find that using absolute velocity as a scalar to form the fifth-dimensional feature of the point cloud improves the detection accuracy. Intuitively, velocity as scalar input can well distinguish between dynamic objects and static environment, and the dynamic targets have high probabilities of being detected.

To step further, we hope to not only input velocity as scalar but also consider the natural direction information of radial velocity and estimate the future distribution of point clouds. To this end, we propose a novel predictive object detection (POD) framework. The main idea of POD is to generate a virtual point cloud of the future frame with the position and velocity of the current frame point and form two frame point clouds. Then the spatiotemporal features are encoded with 4D voxel encoders from the two-frame point cloud. After the voxel encoders, the 4D voxels with features are separated and mapped to current and future BEV features. The current and future BEV features are finally decoded as the detection results of the current frame and the future frame under the current coordinate, respectively. We name the detection of the future frame objects as the predictive object detection task. A bird's eye view (BEV) visualization example of current and predictive object detection of future 0.5 second is shown in Fig. \ref{fig:POD_intro}. 

Despite pre-processing steps for FMCW LiDAR, the key network design of POD is the 4D voxel encoder. A traditional 3D voxel encoder extracts voxel features while maintaining the sparsity of the point cloud. There are two mainstream designs of voxel encoder, Sparse Convolutional VoxelNets\cite{VoxelNet, second} and Voxel Transformers\cite{SST, DSVT}. We extend both paradigms to 4D encoding schemes. The 4D SparseConv VoxelNet achieves higher accuracy on the predictive object task at the expense of higher computation costs. The 4D Voxel Transformer is faster, more efficient, and achieves better performance on the standard detection task.


In summary, our contributions are listed as follows:

\begin{itemize}
\item We explore the unique advantages of FMCW LiDAR in object detection tasks, and propose a novel predictive object detection task with a single frame point cloud.

\item We explore detection network design to better adapt FMCW LiDAR. We design two new modules, virtual point cloud generation and 4D voxel encoder, to make better use of vectorized absolute velocity provided by FMCW LiDAR.

\item We validate the effectiveness of our proposed LiDAR-based predictive detector with our in-house LiDAR dataset. We compare the effectiveness and efficiency of proposed voxel encoders.

\end{itemize}
\section{Related Works}
\subsection{LiDAR-based 3D Object Detection} 
LiDAR-based 3D object detection is a widely investigated task in the field of autonomous driving. VoxelNet\cite{VoxelNet} proposes a general paradigm of a LiDAR-based 3D object detector, which consists of a voxel feature encoding (VFE) layer, a 3D voxel encoder, a BEV encoder and a prediction head. A commonly used solution is to use sparse convolution layers (Spconv) as voxel encoder, ResNet\cite{ResNet} blocks as BEV backbone, anchor head\cite{second} or center head\cite{centerpoint} or query-based head\cite{transfusion} as alternatives of the prediction head. Recent advances in LiDAR detectors focus on the new structures of the voxel encoder. SST\cite{SST} proposes the first single-stride pure sparse Voxel Transformer encoder. FSD series\cite{FSD, FSD++, FSDv2} further improve SST's backbone design and propose a purely sparse prediction head. DSVT\cite{DSVT} applies the normal transformer operators from Pytorch to implement a single-stride 3D voxel backbone, which greatly reduces the deployment difficulty. LION\cite{LION} uses operators such as Mamba\cite{Mamba} to build a multi-stride voxel backbone and improve detection accuracy at the expense of increased inference runtime. Inspired by 4D sparse feature encoding methods such as Flow4D\cite{Flow4D}, our proposed detector further extends the 3D voxel encoders to 4D encodings and finds a balance between detection accuracy and running speed.

\subsection{Prediction with Detection} 
A common perception system consists of 3D object detection, tracking, and prediction, but some works integrate prediction with detection. The major advantage of integrating prediction with detection is that it can deal with possible dangers in driving with a shorter reaction time. Detra\cite{Detra} proposes a unified model to handle object detection and motion forecasting. Motion Inspired Detection\cite{MotionInspiredDetection} proposes to augment offline auto-labeling pipelines with LiDAR scene flow. Predict To Detect (P2D)\cite{PredictToDetect} predicts objects in the current frame using only past frames to learn temporal motion features as a novel prediction-guided temporal aggregation method for vision-BEV methods. Compared to P2D\cite{PredictToDetect}, our proposed detector does not use historical frames while achieving the location of predictive objects in future frames, thus further reducing reaction time.

\subsection{Autonomous Perception with FMCW LiDAR}
FMCW LiDAR is a next-generation 4D LiDAR product compared to the traditional 3D LiDAR as it can provide the radial velocity of each point. The physical properties of FMCW LiDAR make it easy to distinguish between dynamic objects and static backgrounds in the environment. However, the research of FMCW LiDAR in perception is limited due to limited access to latest FMCW LiDAR products. HeLiPR\cite{HeLiPR} proposes a heterogeneous FMCW LiDAR dataset for inter-LiDAR place recognition. DICP\cite{DICP} proposes a Doppler iterative closest point algorithm for point cloud registration of FMCW LiDAR\cite{yoneda2023extended}. Some works\cite{wu2022picking,yoon2023need,zhao2024fmcw,zhao2024free,pang2024efficient} study odometry tasks with FMCW LiDAR as FMCW LiDAR can well classify dynamic objects and static scenes. Gu et al.\cite{gu_iros22} explore object tracking performance with FMCW LiDAR. Given the absence of research on FMCW LiDAR detection tasks, we propose novel detection networks that well match the physical mechanism of FMCW LiDAR.  
\section{Methodology}

\begin{figure*}[ht]
	\centering
\includegraphics[width=1.0\textwidth]{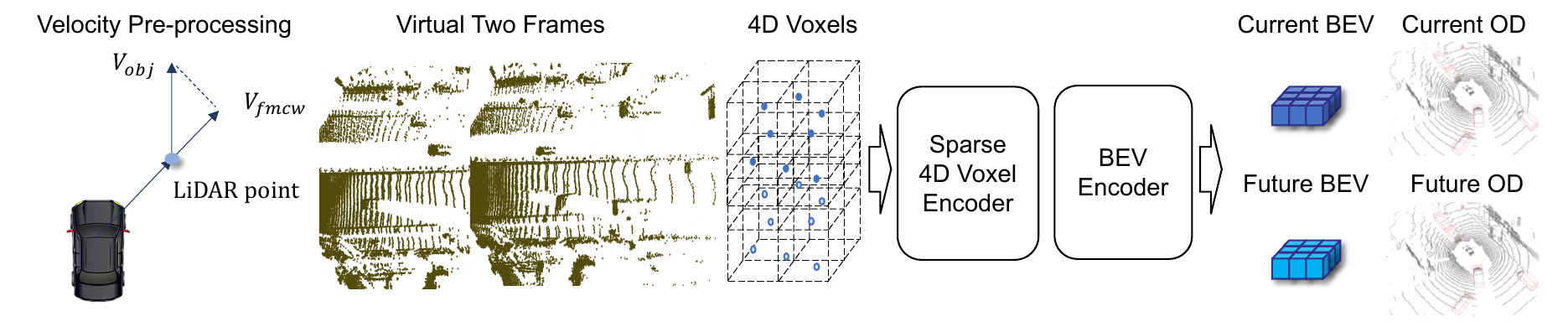}
	\caption{The framework of POD. The framework consists of three important components. (Left): The pre-processing forms virtual two frames and voxelizes point clouds into 4D voxels. (middle): The network encodes 4D voxel features to two-frame BEV features. (Right): Two-frame BEV features are decoded for the current and future detection results.
	}
	\label{fig:POD_framework}
\end{figure*}

\subsection{Preliminaries}
\subsubsection{Sensory input of FMCW LiDAR} The raw data of a single-frame FMCW point cloud is an array with shape $[N, 5]$, where $N$ is the number of points, and the five dimensions are $[x, y, z, i, v]$, where $x, y, z$ are the three-dimensional coordinate position, $i$ is the reflection intensity, and $v$ is the relative radial velocity.

\subsubsection{Formulation of Predictive Object Detection} 

Based on the formulation of the standard object detection task, we propose predictive object detection (POD) as a novel perception task. Given a fixed future horizon, e.g. $0.1s$, $0.2s$, or $0.5s$ starting from the current frame, the model needs to detect the future bounding boxes in the current frame coordinate. For evaluation, the future frame ground-truth boxes are also transformed to the current frame coordinate. The evaluation process is same as standard object detection.

\subsection{Architecture}
The architecture of POD is shown in Fig. \ref{fig:POD_framework}. The network generally follows the design of a standard object detection network, which consists of a voxel feature encoder (VFE), a voxel encoder, a BEV encoder, and a prediction head. The non-learning generation process of future virtual points is handled before the VFE layer. During this process, the single-frame point cloud is transformed to two-frame point clouds. The VFE layer voxelizes two-frame points clouds to 4D voxels. The 4D voxel encoder extracts the spatio-temporal correlation of voxel features while maintaining the voxel indices. After the voxel encoder, 4D voxel indices from two frames are separated and mapped to two-frame BEV features. The two frame BEV features are finally decoded as standard object detection results and predictive object detection results.

\subsection{Virtual Future Points Generation}\label{sec:virtual_point}
\subsubsection{From Relative Velocity to Absolute Velocity}
To utilize the velocity information from FMCW LiDAR, we first pre-process the point cloud to distinguish between dynamic and static objects. Since the ego vehicle is moving, the relative velocity of most points with respect to the vehicle is non-zero. We calculate the absolute velocity by superimposing the relative velocities, which are then used to differentiate between dynamic and static objects. A straightforward approach is to apply a rule-based clustering method to extract the majority of ground points. The average relative velocity of these ground points is then computed, and this average value is subtracted from the velocities of all points. Experimental results demonstrate that this simple rule-based method effectively distinguishes dynamic and static objects.

\subsubsection{Future Point Extrapolation Based on Absolute Velocity}

\begin{figure}[ht]
	\centering
\includegraphics[width=0.35\textwidth]{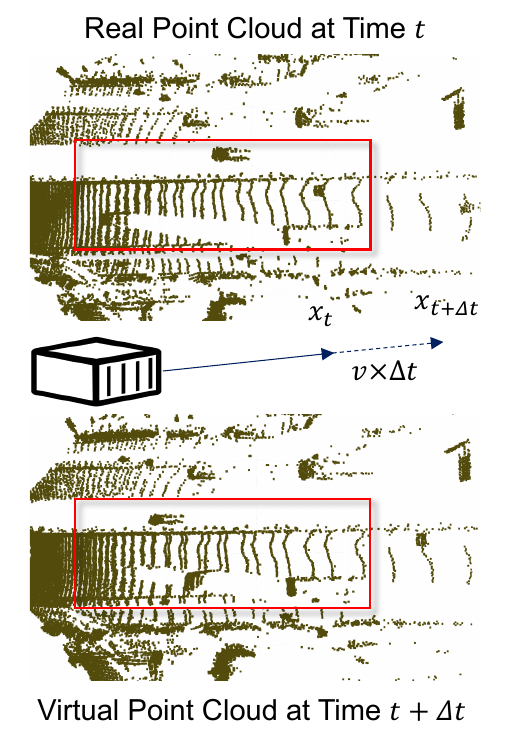}
	\caption{An example of the virtual point generation preprocessing process. The locations of fast-moving vehicles in the current frame and the virtual future frame are highlighted in the red boxes. 
	}
	\label{fig:POD_preprocess}
\end{figure}

Given the absolute radial velocity of the points, we intend to extrapolate the future positions of the point clouds. We parametrize the extrapolation as a ray-tracing process: 

Given a point $x$ from the future LiDAR point cloud $X$, as a ray that starts from the sensor origin $\mathbf o$, travels along the direction $\mathbf d$, and reaches the end point $\mathbf x$ after a distance of $v_{abs} \times \delta t$:
\begin{equation}
    x_{\delta t} = x + v_{abs} \cdot \delta t \cdot \mathbf d, \mathbf x \in \mathbf X
\end{equation}

where $v_{abs}$ is the absolute velocity of point $x$ and $\delta t$ is the future prediction horizon, $x_{\delta t}$ is the extrapolated virtual future point corresponding to the current frame point $x$. 
An example of point cloud before and after the virtual future point generation is shown in Fig. \ref{fig:POD_preprocess}.

\subsection{4D 
Voxel Encoders}\label{sec:4dsparsevoxeltransformer}

After preprocessing of virtual point cloud generation, the VFE layer voxelizes the two-frame point clouds into four-dimensional grid indices $[x, y, z, t]$, where $x$, $y$, $z$ are the indices in the three-dimensional space, and $t$ is the time index. For the input of two-frame point clouds, the value of $t$ index ranges between $0$ or $1$. The VFE layer extracts features within each voxel through a multi-layer perceptron (MLP).

The 4D voxel encoder needs to extract spatiotemporal information within the virtual two frames. Our design target is to ensure both feature extraction efficiency and computational efficiency. Computational efficiency in point cloud encoder design is closely related to the sparse nature of point clouds. Based on existing solutions, we extend two mainstream voxel encoders to fit the 4D voxelized input.

\subsubsection{SparseConv 4D VoxelNet}
\begin{figure}[ht]
	\centering
\includegraphics[width=0.5\textwidth]{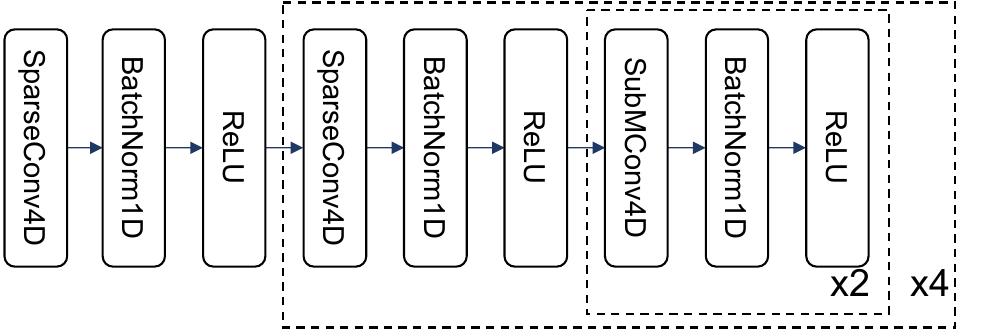}
	\caption{Graphical illustration of SparseConv-based 4D VoxelNet.
	}
	\label{fig:POD_spconv4d}
\end{figure}

We conduct a simple extension to the SparseConv-based VoxelNet. The main idea is to replace the necessary SparseConv3D layers and SubMConv3D layers in the traditional VoxelNet with SparseConv4D layers and SubMConv4D layers\footnote{We use SparseConv4D and SubMConv4D layer from Spconv repository: \url{https://github.com/traveller59/spconv}}, without modifying other parts of the implementation. The network design is shown in Fig. \ref{fig:POD_spconv4d}. Each block consists of one SubMConv4D or SparseConv4D layer, one batch normalization layer and one ReLU activation layer. The network has one input block with SubMConv4D and four cascade downsampling modules. Each downsample module consists of one block with SparseConv4D and two blocks with SubMConv4D. 


\subsubsection{4D Voxel Transformer}

Voxel Transformer is a fully sparse approach and the computational efficiency is ensured by the design of single-stride, non-downsampling feature encoding. There are different implementations of Voxel Transformers, we build our 4D Voxel Transformer upon DSVT, as the voxel attention of DSVT is implemented with standard PyTorch multi-head attention operators. 

\begin{figure}[ht]
	\centering
\includegraphics[width=0.5\textwidth]{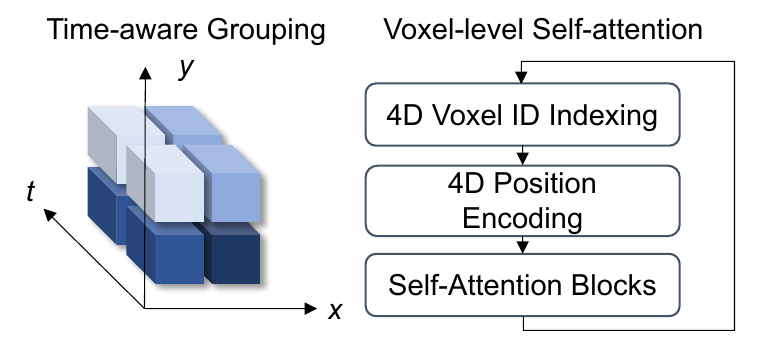}
	\caption{Graphical illustration of 4D Voxel Transformer.
	}
	\label{fig:POD_dsvt4d}
\end{figure}

The graphical illustration of the 4D Voxel Transformer is shown in Fig. \ref{fig:POD_dsvt4d}. The 4D Voxel Transformer has several steps: dynamic window set partition, rotated set intra-window attention, inter-window feature propagation, and 4D pooling. We aim to minimize modifications to the DSVT module design and adapt it at the lowest cost. We find that the core modification is to transform the 3D local window of the Voxel Transformer into a 4D window and align all other operations accordingly. To be specific, temporal-aware 4D voxels are partitioned into non-overlapping 4D windows with size $L \times W \times H \times T$, where $L$, $W$, $H$ and $T$ are the length, width, height, and time dimensions of a 4D window. These 4D windows are further split into 4D window-bounded non-overlapping subsets with a fixed number of voxels. Each voxel index has a learnable 4D positional encoding. The intra-window attention, window shifting, and pooling strategies are the same as DSVT except for minor modifications to feature channels. 

\subsection{Other Modules}

\subsubsection{BEV Encoder}
After 4D voxel encoder, the 4D voxel features are densified. The dense 4D feature is further separated into two 3D features when the time dimension is collapsed. The height dimension of each feature is compressed. We adopt some residual connection layers and multiple down-sampling and up-sampling blocks to encode BEV features at multiple scales.

\subsubsection{Prediction Head}
Our framework is generally flexible with most BEV-based prediction heads. We implement Centerhead from CenterPoint\cite{centerpoint}. The Centerhead predicts the center location of objects along with their attributes including heatmap centerness, location, height, dimension, rotation, and direction.  

\subsubsection{Loss Design}
The losses of both standard and predictive object detection are the same. The losses include heatmap loss, size regression loss, dimension regression loss, and orientation regression loss. We use L1 loss for all regression sub-tasks and cross-entropy loss for classification sub-tasks. The weights for all losses are set as $1.0$.

\section{Experiments}
\subsection{Datasets and Metrics} 
Currently, there is no public dataset and benchmark for FMCW LiDAR object detection, and for data confidentiality, we use the in-house dataset, which is a dataset collected by an experimental 128-line FMCW LiDAR, the farthest detection range of this LiDAR is set as 200 meters, the detection field of view (FOV) is 100 degrees, and the acquisition frequency is 10Hz. We collect street view driving data in Shanghai, China, and other places. The length of each scene is 10 seconds. 32293 frames are selected as training frames and 5,732 frames are selected as testing frames. 

We annotate the collected point clouds both manually and automatically. We label five categories of objects: car, pedestrian, cyclist, van, and traffic cone. We follow the detection metrics proposed from the KITTI\cite{kitti} dataset (AP Recall 40), for which the intersection and union ratio (IoU) thresholds of car and van are 0.5, and the thresholds of the other three categories are 0.25. We show the 3D mAP under Recall 40 of five categories.

For standard object detection, we measure ground-truth boxes with at least one point inside each box. For predictive object detection, we measure all ground-truth boxes annotated by future frame point clouds and transform future bounding boxes to the current coordinate. The ego pose is provided with LiDAR simultaneous localization and mapping (SLAM) algorithms.

\subsection{Implementation Details} 
We build our detection codebase upon OpenPCDet\cite{OpenPCDet} and MMDetection3D\cite{mmdet3d2020} frameworks. We adopt similar implementations of the VFE layer, BEV encoder, and Centerhead from OpenPCDet\cite{OpenPCDet}. 
By default, we use point clouds with five dimensions: x, y, z, intensity, and absolute velocity, as we find that inputting velocity as scalar input is beneficial to the detection task. We apply common training data augmentation techniques including object sampling, random flipping, rotation, and translation. For the predictive object detection task, as the current frame point cloud does not match future bounding boxes, the object sampling augmentation is disabled.

For the setup of SparseConv4D-based detector, the voxel size is $[0.08m,0.08m,0.25m]$ and the grid size is $[1888, 1280, 64]$. The kernel size of each SparseConv layer is $[3,3,3,3]$ and the padding of each SparseConv layer is $[1,1,1,1]$. The downsampling strides of four downsampling modules are $[2,2,2,1]$, $[2,2,2,1]$, $[1,1,2,1]$, and $[1,1,2,1]$ on $x, y, z, t$ dimensions, respectively. 

For the setup of DSVT4D voxel encoder, the backbone is built with one DSVT4D block and two layers in this block. The voxel size is $[0.32m,0.32m,16m]$ and the grid size is $[472, 320, 1]$. By default, we set the window shape as $[60, 60, 1, 2]$, the hybrid factor as $[1,1,1,1]$, and the shifts list as $[[0,0,0,0], [30,30,0,0]]$. The maximum number of subsets is $120$. The input channel is $64$ and the feed-forward channel is $128$.

Each model is trained for 90 epochs with a single A800 GPU and a batch size of 16. The inference time is measured by a single 3090 GPU. The training time and inference speed of different networks are listed in Table. \ref{tab:inference_speed}. The SpConv-based detector has the best efficiency for both training and inference, SpConv4D greatly increases the latency by 75.32\% and training time by 106.81\%. In contrast, DSVT is less efficient than SpConv, but DSVT4D increases marginally as compared to DSVT, with latency by 7.96\% and training time by 14.5\%. DSVT4D is generally more efficient than SpConv4D.

\begin{table}
    \centering
    \caption{Training time and inference speed of different networks. Training time is measured by a single A800 GPU with batch size 16 and 90 epochs for 32k training samples. Inference speed is measured by a single 3090 with batch size 1. }
    \resizebox{0.5\textwidth}{!}{
        \begin{tabular}{l|lll}
            \toprule
            Backbone & Latency (ms) & FPS & Training Time (GPU hour) \\
            \midrule
            SpConv   & 54.3 & 18.4  & 44 \\
            SpConv4D (ours)  & 95.2 & 10.5 & 91  \\
            DSVT    & 74.1 & 13.5 & 55  \\
            DSVT4D (ours) & 80.0 & 12.5 & 63  \\
            \bottomrule
        \end{tabular}
    }

    \label{tab:inference_speed}
\end{table}

\subsection{Results on Standard Object Detection}

The results of standard object detection are shown in Table. \ref{tab:standard_od}. We first implement two well-established methods, CenterPoint\cite{centerpoint} and DSVT\cite{DSVT} on FMCW LiDAR. We find that with careful hyperparameter setting, DSVT outperforms CenterPoint in all categories. 
We implement SpConv4D as the extension of SpConv. SpConv4D shows better precision on the traffic cone, which is $3.03$ higher than SpConv, but for other free categories, SpConv4D does not work as well as SpConv. 

We implement two extensions of DSVT. A naive solution is to concatenate the two frames together, encode them with the DSVT 3D voxel encoder and retain voxel features that belong to the current frame voxel coordinates. The cat2frame solution has inferior performance compared to the original DSVT implementation in all categories. 

We implement DSVT4D as a successful 4D extension of DSVT, which shows the best precision in terms of car $mAP=90.34$, pedestrian $mAP=66.15$, and traffic cone categories $mAP=76.56$.

\begin{table*}
    \centering
    \caption{Standard object detection results of detectors with different voxel encoders. CenterPoint\cite{centerpoint} and DSVT\cite{DSVT} are adapted from open-source code with carefully tuned hyperparameters. All methods use the same CenterHead. 'cat2frame' refers to the baseline that concatenates two frames. T. C. refers to the traffic cone.}
    \resizebox{1.0\textwidth}{!}{
        \begin{tabular}{ll|lllll}
            \toprule
            Method       & Voxel Encoder               & $mAP_{3D}@Car$ & $mAP_{3D}@Pedestrian$ & $mAP_{3D}@Cyclist$ & $mAP_{3D}@Van$ & $mAP_{3D}@T.C.$ \\
            \midrule
            Centerpoint  & SpConv                & 89.17 & 61.62 & 85.44 & 64.86 & 69.04 \\
            DSVT         & DSVT-Pillar         & 89.40 & 65.62 & \textbf{86.33} & \textbf{68.63} & 72.54 \\
            POD (ours) & DSVT-Pillar-cat2frame & 88.41 & 55.77 & 80.11 & 60.64 & 63.61 \\
            
            POD (ours)& SpConv4D   & 86.91 & 60.44 & 84.10 & 57.47 & 72.07 \\
        
            POD (ours)& DSVT4D              & \textbf{90.34} & \textbf{66.15} & 85.41 & 66.65 & \textbf{76.56} \\
            \bottomrule
        \end{tabular}
    }

    \label{tab:standard_od}
\end{table*}

\subsection{Results on Predictive Object Detection}
To benchmark the models' performance on predictive object detection task, we set the future prediction horizons of $0.1s$, $0.2s$ and $0.5s$. The results are shown in Table. \ref{tab:predictive_od}. When the predictive horizon is short, the detector achieves reasonable accuracy, and when the predicted time domain is extended, the accuracy decreases rapidly. The main reason for rapid decrease in precision is that using the IoU-based detection metric for prediction is very strict. On the other hand, the single-frame radial velocity information only indicates the instantaneous state, so the network mostly predicts the future dynamics of objects based on inertia. For the prediction horizon of 0.1s, DSVT4D performs better on vehicles (car and van) than the SpConv4D detector. For longer prediction horizons, SpConv4D outperforms DSVT4D remarkably in all categories at the expense of higher computation costs. 

We further train both models on a 6k subset of the 32k training set and report the predictive object detection capabilities of the models when trained with a small amount of data. We find that on the 6k subset, the detection precision of the models for major classes, such as car, can sometimes even exceed that of the whole set training, but the models trained with the whole set generally have higher precision in the sum of all categories and are more balanced in all categories. The efficient training of the subset proves that the proposed models have considerable training efficiency that can achieve a good prediction ability with fewer labeled data.
\begin{table*}
    \centering
    \caption{Predictive object detection results of detectors with different voxel encoders. We train the predictive detection for two models, DSVT4D and SpConv4D, on two data scales: 6k and 32k training samples, and on three future prediction horizons: 0.1s, 0.2s and 0.5s. T. C. refers to the traffic cone.}
    \resizebox{1.0\textwidth}{!}{
        \begin{tabular}{lll|lllll}
            \toprule
            Encoder & Horizon & Data  & $mAP_{3D}@Car$ & $mAP_{3D}@Pedestrian$ & $mAP_{3D}@Cyclist$ & $mAP_{3D}@Van$ & $mAP_{3D}@T.C.$ \\
            \midrule
            DSVT4D & 0.1s   & 6k & \textbf{47.23} & 37.46 & 57.79 & 34.55 & 59.44 \\

            SpConv4D & 0.1s & 6k & 46.08 & 32.49 & 54.84 & 34.58 & 63.62 \\

            DSVT4D & 0.1s   & 32k   & 45.35 & 43.40 & 57.29 & \textbf{39.68} & 57.91 \\
            SpConv4D & 0.1s & 32k & 43.44 & \textbf{47.77} &  \textbf{60.26} & 35.96 & \textbf{65.10} \\\midrule

            DSVT4D & 0.2s   & 6k & 31.21 & 21.08 & 25.06 & 19.09 & 44.93 \\

            SpConv4D & 0.2s & 6k & 34.35 & 14.58 & 30.72 & 20.57 & \textbf{53.45}\\

            DSVT4D & 0.2s  & 32k  & 30.66 & 20.48 & 21.09 & 24.11 & 45.70 \\
            SpConv4D & 0.2s & 32k  & \textbf{35.66} & \textbf{23.95} & \textbf{36.36} & \textbf{25.08} & 53.09 \\\midrule

            DSVT4D & 0.5s & 6k & 20.99 & 1.91 & 10.22 & 8.79 & 23.19\\

            SpConv4D & 0.5s & 6k & 22.41 & 1.97 & 12.43 & 7.60 & 30.43\\

            DSVT4D & 0.5s & 32k    & 21.79 & 5.16 & 14.82 & 8.81 & 26.18 \\
            SpConv4D & 0.5s  & 32k    & \textbf{22.78} & \textbf{7.49} & \textbf{17.99} & \textbf{17.66} & \textbf{31.76} \\
            \bottomrule
        \end{tabular}
    }

    \label{tab:predictive_od}
\end{table*}

\subsection{Qualitative Results}
\begin{figure*}[ht]
	\centering
\includegraphics[width=1.0\textwidth]{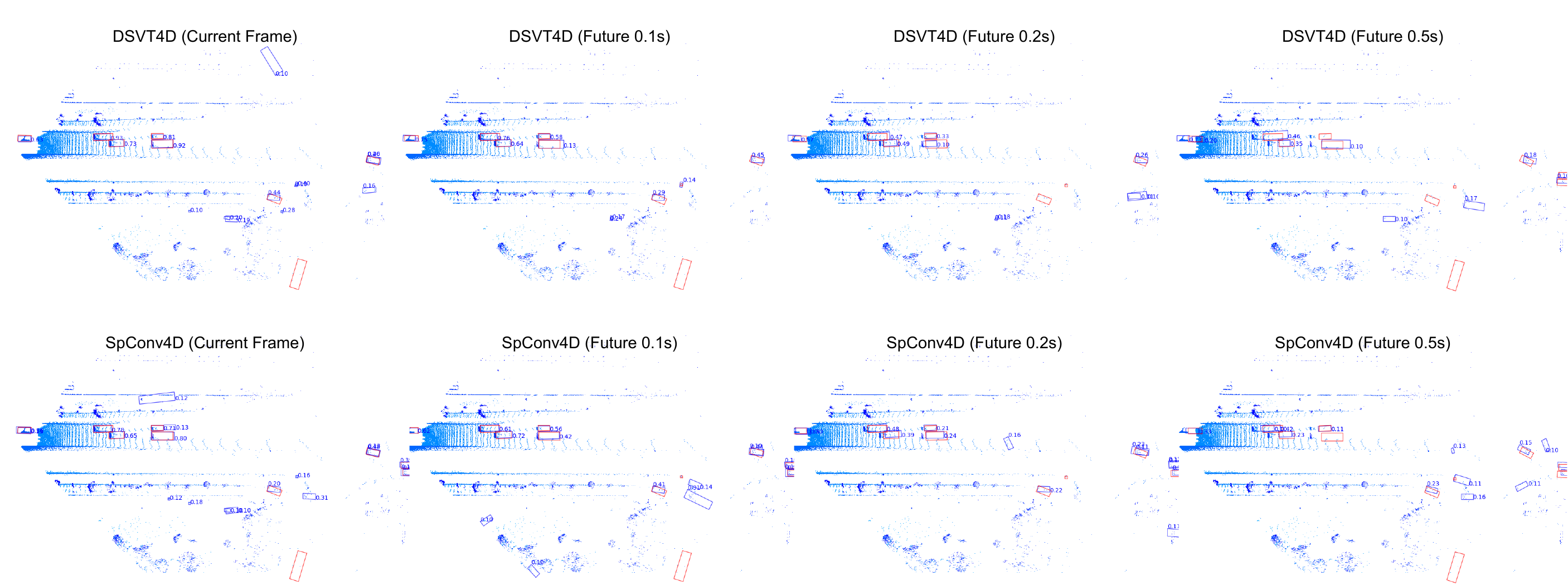}
	\caption{The visualizations of POD. In this driving scene, the ego vehicle is driving straight with several vehicles driving ahead. Boxes in blue are predictions and boxes in red are ground-truth. 
	}
	\label{fig:POD_visualizations}
\end{figure*}

A single-frame visualization of different networks and different prediction horizons is shown in Fig. \ref{fig:POD_visualizations}. Both methods well solve the current frame detection. As the prediction horizon advances, SpConv4D better detects the future bounding boxes, especially for vehicles ahead of the ego vehicle. 
\begin{table}
    \centering
    \caption{Ablations on how different inputs affect detection performance. $x, y, z, intensity$ is the location and intensity of point clouds. relv and absv refer to relative velocity and absolute velocity.}
    \resizebox{0.5\textwidth}{!}{
        \begin{tabular}{l|lllll}
            \toprule
            Input  & Car & Pedestrian & Cyclist & Van & T.C. \\
            \midrule
            xyz     & 87.03 & 65.56 & 79.77 & 59.47 & 62.65 \\
            xyzi     & 87.62 & 65.79 & 82.37 & \textbf{60.89} & 67.07 \\
            xyzi\_relv    & 80.46 & 52.84 & 79.73 & 48.49 & 42.76 \\
            xyzi\_absv   & \textbf{87.89} & \textbf{73.99} & \textbf{83.97} & 58.80 & \textbf{67.15} \\

            \bottomrule
        \end{tabular}
    }

    \label{tab:ablate_vel}
\end{table}
The visualization indicates that both detectors have a tendency to overestimate the box dimension (length and width) of future objects, which is caused by the mechanism that the extrapolation of points makes the future distribution dispersed, but this tendency will not affect the safety of the driving space.

\subsection{Ablation Study}
We conduct ablative experiments to demonstrate the efficacy of FMCW LiDAR raw data and model designs.

\subsubsection{Effects of Different Velocity Preprocessing on Detection Task}
We show the effects of velocity as input on CenterPoint model with SpConv backbone on standard object detection task in Table. \ref{tab:ablate_vel}. Point cloud input $x, y, z, i$ with intensity improves precision in all categories compared to point cloud without intensity. Using relative velocity as input severely deteriorates the detection accuracy, as relative velocity is highly dependent on ego vehicle velocity. Using absolute velocity as input improves the accuracy for most obstacles. The introduction of absolute velocity greatly enhances the detection accuracy of pedestrians to $8.2 mAP$, as intuitively upright objects with velocity can be distinguished as pedestrians.

\subsubsection{Effects of Network Parameter Designs in 4D Voxel Encoder}
We show the effects of different network parameter designs for the predictive object detection task in Table. \ref{tab:ablate_net}. We set different window shapes, shifts, and sets. The results show that a larger window shape with larger shifts has better precision on large obstacles, such as cars and vans, while a smaller window shape with smaller shifts has better precision of small obstacles, including pedestrians and traffic cones.

\begin{table*}
    \centering
    \caption{Ablations on Network designs of 4D Voxel Transformer.}
    \resizebox{1.0\textwidth}{!}{
        \begin{tabular}{lll|lllll}
            \toprule
            Window Shape & Shift & Set Num. & $mAP_{3D}@Car$ & $mAP_{3D}@Pedestrian$ & $mAP_{3D}@Cyclist$ & $mAP_{3D}@Van$ & $mAP_{3D}@T.C.$ \\
            \midrule
            $[60,60,1,2]$  & $[30,30,0,0]$ & 120  & \textbf{47.23} & 37.46 & 57.79 & 34.55 & 58.44 \\
            $[30,30,1,2]$  & $[15,15,0,0]$ & 90 & 39.78 & 43.89 & 57.61 & 28.95 & 58.43 \\    
            $[30,30,1,2]$  & $[15,15,0,0]$ & 120 & 42.24 & \textbf{45.03} & 57.97 & 33.56 & \textbf{59.97}\\    
            $[90,90,1,2]$  & $[45,45,0,0]$ & 120 & 46.44 & 43.80 & \textbf{58.82} & \textbf{37.91} & 57.68\\              
            \bottomrule
        \end{tabular}
    }

    \label{tab:ablate_net}
\end{table*}
\subsection{Discussion and Limitations}
In our experiments, we verify that POD has the ability to conduct predictive object detection in a single frame, but there are still some limitations that need to be discussed. In terms of data, we do not collect enough transient hazard events in our dataset to verify the detection performance of the detector in partially occluded conditions. 

In terms of model, we do not consider the problem of missing tangential velocity in the FMCW LiDAR in model design, and only used the radial velocity to extrapolate the future virtual point cloud. This may lead to failed predictions of tangentially moving objects, such as a pedestrian straight in front of a vehicle crossing a pedestrian crossing. The utilization of more historical frames allows for a better estimate of tangential velocity and avoids this failure. We leave these two points for future work.

In terms of training strategies, we train standard and predictive object detection separately, as object sampling data augmentation improves current-frame detection, but it is not applicable to predictive detection, as future box labels are not attached to current points. Moreover, we only support one fixed prediction horizon per model training at the current stage. Joint training of current and flexible future frames remains a future work.

\section{Conclusion}
This paper demonstrates the first practice of future predictive object detection (POD) with single-frame FMCW point clouds. The idea of virtual point generation is straightforward and the proposed voxel encoders are highly effective and efficient in encoding virtual two-frame voxel features. Qualitative and quantitative experiments with our in-house FMCW LiDAR dataset show competitive performance and great potential in the perception system of autonomous driving. For future work, we hope to explore the key role of single-frame prediction in avoiding transient risks. We will collect data from transient-hazard test scenarios such as ghost probes, test on critical data, and open-source the dataset collected with FMCW LiDAR under the premise of regulatory requirements.


\bibliographystyle{IEEEtran}
\bibliography{ref}

\begin{thebibliography}{10}
\providecommand{\url}[1]{#1}
\csname url@rmstyle\endcsname
\providecommand{\newblock}{\relax}
\providecommand{\bibinfo}[2]{#2}
\providecommand\BIBentrySTDinterwordspacing{\spaceskip=0pt\relax}
\providecommand\BIBentryALTinterwordstretchfactor{4}
\providecommand\BIBentryALTinterwordspacing{\spaceskip=\fontdimen2\font plus
\BIBentryALTinterwordstretchfactor\fontdimen3\font minus \fontdimen4\font\relax}
\providecommand\BIBforeignlanguage[2]{{%
\expandafter\ifx\csname l@#1\endcsname\relax
\typeout{** WARNING: IEEEtran.bst: No hyphenation pattern has been}%
\typeout{** loaded for the language `#1'. Using the pattern for}%
\typeout{** the default language instead.}%
\else
\language=\csname l@#1\endcsname
\fi
#2}}

\bibitem{velovox}
T.~Ma, Z.~Zheng, H.~Zhou, X.~Cai, X.~Yang, Y.~Li, B.~Shi, and H.~Li, ``Velovox: A low-cost and accurate 4d object detector with single-frame point cloud of livox lidar,'' in \emph{2024 IEEE International Conference on Robotics and Automation (ICRA)}.\hskip 1em plus 0.5em minus 0.4em\relax IEEE, 2024, pp. 1992--1998.

\bibitem{VoxelNet}
Y.~Zhou and O.~Tuzel, ``Voxelnet: End-to-end learning for point cloud based 3d object detection,'' in \emph{Proceedings of the IEEE conference on computer vision and pattern recognition}, 2018, pp. 4490--4499.

\bibitem{second}
Y.~Yan, Y.~Mao, and B.~Li, ``Second: Sparsely embedded convolutional detection,'' \emph{Sensors}, vol.~18, no.~10, p. 3337, 2018.

\bibitem{SST}
L.~Fan, Z.~Pang, T.~Zhang, Y.-X. Wang, H.~Zhao, F.~Wang, N.~Wang, and Z.~Zhang, ``Embracing single stride 3d object detector with sparse transformer,'' in \emph{Proceedings of the IEEE/CVF conference on computer vision and pattern recognition}, 2022, pp. 8458--8468.

\bibitem{DSVT}
H.~Wang, C.~Shi, S.~Shi, M.~Lei, S.~Wang, D.~He, B.~Schiele, and L.~Wang, ``Dsvt: Dynamic sparse voxel transformer with rotated sets,'' in \emph{Proceedings of the IEEE/CVF Conference on Computer Vision and Pattern Recognition}, 2023, pp. 13\,520--13\,529.

\bibitem{ResNet}
K.~He, X.~Zhang, S.~Ren, and J.~Sun, ``Deep residual learning for image recognition,'' in \emph{Proceedings of the IEEE conference on computer vision and pattern recognition}, 2016, pp. 770--778.

\bibitem{centerpoint}
T.~Yin, X.~Zhou, and P.~Krahenbuhl, ``Center-based 3d object detection and tracking,'' in \emph{Proceedings of the IEEE/CVF conference on computer vision and pattern recognition}, 2021, pp. 11\,784--11\,793.

\bibitem{transfusion}
X.~Bai, Z.~Hu, X.~Zhu, Q.~Huang, Y.~Chen, H.~Fu, and C.-L. Tai, ``Transfusion: Robust lidar-camera fusion for 3d object detection with transformers,'' in \emph{Proceedings of the IEEE/CVF conference on computer vision and pattern recognition}, 2022, pp. 1090--1099.

\bibitem{FSD}
L.~Fan, F.~Wang, N.~Wang, and Z.-X. Zhang, ``Fully sparse 3d object detection,'' \emph{Advances in Neural Information Processing Systems}, vol.~35, pp. 351--363, 2022.

\bibitem{FSD++}
L.~Fan, Y.~Yang, F.~Wang, N.~Wang, and Z.~Zhang, ``Super sparse 3d object detection,'' \emph{IEEE transactions on pattern analysis and machine intelligence}, vol.~45, no.~10, pp. 12\,490--12\,505, 2023.

\bibitem{FSDv2}
L.~Fan, F.~Wang, N.~Wang, and Z.~Zhang, ``Fsd v2: Improving fully sparse 3d object detection with virtual voxels,'' \emph{IEEE Transactions on Pattern Analysis and Machine Intelligence}, 2024.

\bibitem{LION}
Z.~Liu, J.~Hou, X.~Wang, X.~Ye, J.~Wang, H.~Zhao, and X.~Bai, ``Lion: Linear group rnn for 3d object detection in point clouds,'' in \emph{Advances in Neural Information Processing Systems}, vol.~37, 2025, pp. 13\,601--13\,626.

\bibitem{Mamba}
T.~Dao and A.~Gu, ``Transformers are {SSM}s: Generalized models and efficient algorithms through structured state space duality,'' in \emph{International Conference on Machine Learning (ICML)}, 2024.

\bibitem{Flow4D}
J.~Kim, J.~Woo, U.~Shin, J.~Oh, and S.~Im, ``Flow4d: Leveraging 4d voxel network for lidar scene flow estimation,'' \emph{IEEE Robotics and Automation Letters}, 2025.

\bibitem{Detra}
S.~Casas, B.~Agro, J.~Mao, T.~Gilles, A.~Cui, T.~Li, and R.~Urtasun, ``Detra: A unified model for object detection and trajectory forecasting,'' in \emph{Computer Vision -- ECCV 2024}, A.~Leonardis, E.~Ricci, S.~Roth, O.~Russakovsky, T.~Sattler, and G.~Varol, Eds.\hskip 1em plus 0.5em minus 0.4em\relax Cham: Springer Nature Switzerland, 2025, pp. 326--342.

\bibitem{MotionInspiredDetection}
M.~Najibi, J.~Ji, Y.~Zhou, C.~R. Qi, X.~Yan, S.~Ettinger, and D.~Anguelov, ``Motion inspired unsupervised perception and prediction in autonomous driving,'' in \emph{Computer Vision -- ECCV 2022}, S.~Avidan, G.~Brostow, M.~Ciss{\'e}, G.~M. Farinella, and T.~Hassner, Eds.\hskip 1em plus 0.5em minus 0.4em\relax Cham: Springer Nature Switzerland, 2022, pp. 424--443.

\bibitem{PredictToDetect}
S.~Kim, Y.~Kim, I.-J. Lee, and D.~Kum, ``Predict to detect: Prediction-guided 3d object detection using sequential images,'' in \emph{2023 IEEE/CVF International Conference on Computer Vision (ICCV)}, 2023, pp. 18\,011--18\,020.

\bibitem{HeLiPR}
M.~Jung, W.~Yang, D.~Lee, H.~Gil, G.~Kim, and A.~Kim, ``Helipr: Heterogeneous lidar dataset for inter-lidar place recognition under spatiotemporal variations,'' \emph{The International Journal of Robotics Research}, vol.~43, no.~12, pp. 1867--1883, 2024.

\bibitem{DICP}
B.~Hexsel, H.~Vhavle, and Y.~Chen, ``{DICP: Doppler Iterative Closest Point Algorithm},'' in \emph{Proceedings of Robotics: Science and Systems}, New York City, NY, USA, June 2022.

\bibitem{yoneda2023extended}
M.~Yoneda, K.-M. Dahl{\'e}n, and T.~Ogawa, ``Extended object tracking with doppler velocity-based point registration,'' in \emph{2023 IEEE Symposium Sensor Data Fusion and International Conference on Multisensor Fusion and Integration (SDF-MFI)}.\hskip 1em plus 0.5em minus 0.4em\relax IEEE, 2023, pp. 1--8.

\bibitem{wu2022picking}
Y.~Wu, D.~J. Yoon, K.~Burnett, S.~Kammel, Y.~Chen, H.~Vhavle, and T.~D. Barfoot, ``Picking up speed: Continuous-time lidar-only odometry using doppler velocity measurements,'' \emph{IEEE Robotics and Automation Letters}, vol.~8, no.~1, pp. 264--271, 2022.

\bibitem{yoon2023need}
D.~J. Yoon, K.~Burnett, J.~Laconte, Y.~Chen, H.~Vhavle, S.~Kammel, J.~Reuther, and T.~D. Barfoot, ``Need for speed: Fast correspondence-free lidar-inertial odometry using doppler velocity,'' in \emph{2023 IEEE/RSJ International Conference on Intelligent Robots and Systems (IROS)}.\hskip 1em plus 0.5em minus 0.4em\relax IEEE, 2023, pp. 5304--5310.

\bibitem{zhao2024fmcw}
M.~Zhao, J.~Wang, T.~Gao, C.~Xu, and H.~Kong, ``Fmcw-lio: A doppler lidar-inertial odometry,'' \emph{IEEE Robotics and Automation Letters}, 2024.

\bibitem{zhao2024free}
------, ``Free-init: Scan-free, motion-free, and correspondence-free initialization for doppler lidar-inertial systems,'' \emph{IEEE Robotics and Automation Letters}, 2024.

\bibitem{pang2024efficient}
C.~Pang, Z.~Shen, R.~Wu, and Z.~Fang, ``Efficient doppler lidar odometry using scan slicing and vehicle kinematics,'' \emph{IEEE Transactions on Instrumentation and Measurement}, 2024.

\bibitem{gu_iros22}
Y.~Gu, H.~Cheng, K.~Wang, D.~Dou, C.~Xu, and H.~Kong, ``Learning moving-object tracking with fmcw lidar,'' in \emph{2022 IEEE/RSJ International Conference on Intelligent Robots and Systems (IROS)}, 2022, pp. 3747--3753.

\bibitem{kitti}
A.~Geiger, P.~Lenz, and R.~Urtasun, ``Are we ready for autonomous driving? the kitti vision benchmark suite,'' in \emph{2012 IEEE Conference on Computer Vision and Pattern Recognition}, 2012, pp. 3354--3361.

\bibitem{OpenPCDet}
O.~D. Team, ``Openpcdet: An open-source toolbox for 3d object detection from point clouds,'' \url{https://github.com/open-mmlab/OpenPCDet}, 2020.

\bibitem{mmdet3d2020}
M.~Contributors, ``{MMDetection3D: OpenMMLab} next-generation platform for general {3D} object detection,'' \url{https://github.com/open-mmlab/mmdetection3d}, 2020.

\end{thebibliography}

\end{document}